\documentclass{article}
\usepackage{xcolor}
\usepackage{ijcai26}

\usepackage{times}
\usepackage{soul}
\usepackage{url}
\usepackage[hidelinks]{hyperref}
\usepackage[utf8]{inputenc}
\usepackage[small]{caption}
\usepackage{graphicx}
\usepackage{multirow}
\usepackage{amsmath}
\usepackage{amsthm}
\usepackage{amssymb}
\usepackage{booktabs}
\usepackage{algorithm}
\usepackage{algorithmic}
\usepackage[switch]{lineno}
\usepackage{makecell}   
\usepackage{bm}         

\title{Joint Medical Image Enhancement and Segmentation with Diffusion-based Symbiotic Information Interaction}

\author{
Ying Chen$^1$
\and
Jinyue Li$^2$
\and
Qiankun Li$^{3\dagger}$\\
\affiliations
$^1$Shenzhen Research Institute, The Chinese University of Hong Kong\\
$^2$University of Science and Technology of China\\
$^3$Imperial Global Singapore, Imperial College London\\
\emails
q.li2@imperial.ac.uk
}

\begin{document}
\maketitle
\footnotetext[1]{$\text{Corresponding author}^{\dagger}$ \textit{(q.li2@imperial.ac.uk)}.}

\begin{abstract}
Image quality is critical for accurate medical diagnosis. However, MRI, CT, and ultrasound images are often of low resolution and quality due to cost constraints, complicating the visualization of key anatomical structures and lesions. While such limitations are common in practice, traditional methods treat image enhancement as a separate preprocessing step, failing to fully leverage its potential synergy with image segmentation. To address this, we propose DiSIINet (Diffusion-based Symbiotic Information Interaction Network), which is built on the principle that enhancement and segmentation should mutually reinforce each other in a unified model. Based on Denoising Diffusion Implicit Models (DDIM), DiSIINet integrates an enhancement branch and a segmentation branch. These branches interact through a novel Symbiotic Information Interaction (SII) module, which facilitates dynamic, feature-level information exchange via cross-attention during the reverse diffusion process. This design enables both tasks to iteratively improve each other. The DDIM backbone ensures high-quality output and efficient inference through deterministic sampling. Experiments on multi-modal medical datasets (MRI, CT, ultrasound) show that DiSIINet achieves significant performance improvements compared to sequential or independent enhancement and segmentation approaches. The code is available at: \url{https://github.com/Reconsider80/DiSIINet}.
\end{abstract}

\section{Introduction}

Medical images acquired under suboptimal conditions may encounter quality issues, including blurriness, poor lighting, low resolution, and noise, which can result in misdiagnosis. While advanced medical image enhancement techniques often face challenges in improving high-resolution image quality while preserving clear local anatomical features. Traditional image enhancement methods, particularly those based on transformers, typically operate under the assumption of known degradation types, which often leads to overly smoothed results. StillGAN ~\cite{ma2021structure} introduces an innovative and versatile bi-directional GAN for enhancing medical image quality. Although there have been significant advancements in improving visual perception, methods based on generative adversarial networks (GANs) ~\cite{liang2022efficient,park2023content} still encounter challenges in balancing perceptual quality with fidelity, often causing artifacts. 
 
Recently, the emergence of diffusion models (DMs) ~\cite{ho2020denoising} has demonstrated remarkable capabilities in approximating intricate distributions and generating realistic images. Stable Diffusion ~\cite{rombach2022high} is a conditional diffusion model, whose core concept lies in being a generation system centrally guided by conditional input. Among diffusion-based super-resolution techniques, StableSR ~\cite{wang2023exploring} employs the latent representation of low-quality (LQ) images as the guiding condition for StableDiffusion to achieve super-resolution. In contrast, DiffBIR ~\cite{lin2024diffbir} first restores LQ images before using generative priors to maintain a balance between quality and fidelity during the diffusion process. These approaches highlight the significant potential of generative priors in SR tasks; however, relying solely on LQ image data without additional semantic control may lead to inaccurate content reconstruction. Given the inherent advantages of textual prompts in directing generation within pretrained text-to-image models, recent approaches have begun to utilize semantic descriptions extracted from LQ images to enhance control over the generation process and to improve the semantic accuracy of image restoration. For example, PASD ~\cite{yang2024pixel} and SeeSR ~\cite{wu2024seesr} leverage pretrained captioning or tagging models to derive image content, using textual descriptions as prompts to guide generation. However, these methods may encounter two primary limitations. First, they can occasionally overlook significant components include key objects, textures, or regions that contribute to the overall semantic understanding of the image, resulting in inaccurate semantic details in those areas. Second, a single region may respond strongly to multiple prompts, creating semantic ambiguity in the context of image super-resolution. 

\begin{figure*}[h]
\centering
\includegraphics[scale=0.56]{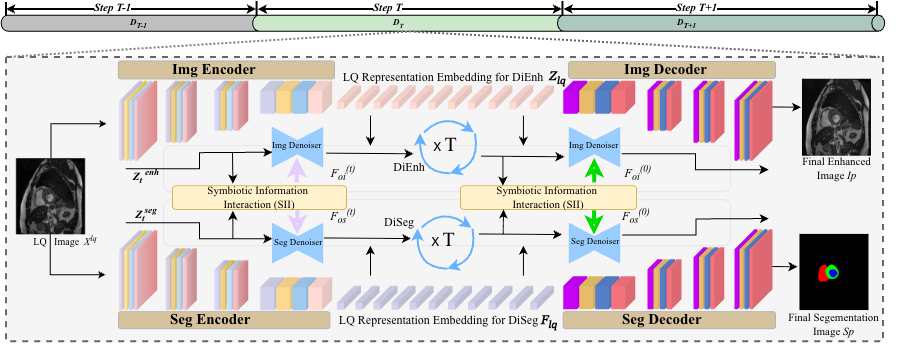}
\caption{The framework of DiSIINet comprises three key parts: (a)DiEnh branch performs image enhancement diffusion process; (b)DiSeg branch conducts semantic segmentation diffusion process; (c)DiEnh and DiSeg collaborate through SII module to achieving mutual benefits. } \label{fig1}
\end{figure*}

In this paper, we aim to address these challenges by utilizing semantic segmentation as an additional control condition. Unlike textual prompts, semantic segmentation provides a more detailed understanding of prominent objects within an image by assigning class labels to individual pixels. However, predicting segmentation masks from heavily degraded images presents significant challenges, and inaccurate masks can lead to erroneous semantic content and distorted spatial structures in the restoration output. Motivated by the fact that image enhancement and segmentation can benefit each other, we propose the joint medical image enhancement and segmentation with diffusion-based Symbiotic Information Interaction (DiSIINet) framework as Fig.\ref{fig1} which introduces a dual diffusion framework to facilitate interaction between the medical image enhancement and segmentation these two closely related tasks. 

The key contributions can be summarized as follows:

\begin{itemize}
\item We propose DiSIINet, the first leverages a dual-branch diffusion architecture where enhancement and segmentation mutually reinforce each other through continuous information exchange.

\item We design an innovative SII module by computing cross-attention between branch features that enables bidirectional information flow between enhancement (DiEnh) and segmentation (DiSeg) branches during reverse diffusion.

\item Extensive experiments demonstrate that our method consistently improves performance across diverse mainstream medical imaging modalities (MRI, CT, ultrasound) and multiple datasets.
\end{itemize}

\section{Related Work}

\subsection{Joint Medical Image Enhancement and Segmentation}

Traditionally, research has addressed these image enhancement and segmentation task independently, but recent efforts ~\cite{xiao2024semantic} explore their synergistic integration. Existing joint approaches can be categorized into two main paradigms: segmentation-guided enhancement and enhancement-guided segmentation.
Segmentation-guided enhancement methods utilize segmentation information to guide the denoising process while preserving structural integrity. Liu et al.~\cite{liu2017image} proposed a deep neural network framework that integrates image denoising with high-level tasks, using a joint loss function to propagate semantic information to low-level visual processing. More recently, the SDABN framework~\cite{xu2023synergy} employs segmentation and denoising blocks (SDBs) in a cascading manner, where each block's segmentation probability map refines subsequent denoising. However, these methods typically establish unidirectional information flow, limiting their ability to fully exploit the mutual benefits between tasks. Enhancement-guided segmentation focuses on improving segmentation robustness by incorporating noise resilience during training. DenoiSeg~\cite{buchholz2020denoiseg} achieved simultaneous segmentation and denoising within a UNet architecture through a combined self-supervised denoising loss. Recent studies in medical imaging have begun to explore deeper synergies. For instance, Wang et al.~\cite{wang2024joint} proposed an Instance-aware Embedding Module (IEM) to leverage segmentation features for denoising, though they did not utilize denoising features to improve segmentation. 

However, There remains a significant gap in methods that facilitate deep feature interaction between two tasks, particularly within the powerful generative framework of diffusion models. Our DiSIINet framework addresses this gap by introducing a symbiotic information interaction mechanism that enables continuous bidirectional information flow between enhancement and segmentation during the diffusion process.

\subsection{Diffusion Models for Medical Image Analysis}

Diffusion models have shown promise in medical image segmentation by leveraging their generative capabilities. Scaling Vision Transformers~\cite{zhai2022scaling} employ stochastic sampling to create an implicit ensemble of segmentations, but suffer from convergence issues and require numerous time-intensive iterations. MedSegDiff-V2~\cite{wu2024medsegdiff} introduces dual conditioning mechanisms—Anchor Condition and Semantic Condition—along with SS-Former to integrate segmentation features into the diffusion process. However, these approaches treat segmentation as an isolated task without considering its potential synergy with image enhancement. Medical image enhancement faces unique challenges due to acquisition-related noise, artifacts, and low signal-to-noise ratios (SNR). Diffusion models have been adapted for various denoising applications. For MRI, $\mathrm{DDM}^2$~\cite{xiang2023ddm} incorporates self-supervised denoising into diffusion models. For low-dose CT (LDCT), CoreDiff~\cite{gao2023corediff} replaces random Gaussian noise with LDCT images as initial states and introduces a mean-preserving degradation operator to reduce sampling steps while preventing structural distortion. 

Despite these advancements, when conditioning is employed such as using segmentation to guide enhancement or vice versa, it is often unidirectional failing to establish symbiotic relationships where both tasks mutually benefit. Our work addresses these gaps by proposing DiSIINet, which integrates enhancement and segmentation within a unified DDIM framework with bidirectional information exchange, enabling mutual reinforcement while maintaining efficient sampling.

\section{Methodology}

\subsection{Dual-Diffusion Joint Medical Image Enhancement and Segmentation}

This framework includes two branches, the diffusion-based image enhancement branch (DiEnh) and the diffusion-based segmentation branch (DiSeg), which are linked by the Symbiotic Information Interaction (SII) module. Inspired by DDIMs \cite{song2021denoising}, which is an efficient variant of DDPMs with accelerated sampling and deterministic generation, DiEnh executes both forward and reverse diffusion processes within the latent space utilizing a pre-trained Variational Autoencoder (VAE) \cite{rombach2022high}, specifically the Image Encoder and Image Decoder components. DDIM retains the same forward process as DDPM but introduces a non-Markovian reverse process that enables accelerated sampling. 

In the DiEnh branch, we leverage the DDIM framework for efficient image enhancement. The branch operates in two phases: training and inference.

During training, we have paired data consisting of low-quality (LQ) images $\mathbf{X}^{lq}$ and their corresponding ground truth (GT) images $\mathbf{I}_{gt}$. We first encode both images into latent space using a VAE encoder:

\begin{equation}
\mathbf{Z}_{lq} = \text{Image Encoder}(\mathbf{X}^{lq}).
\end{equation}

\begin{equation}
\mathbf{Z}^{gt,enh} = \text{Encoder}(\mathbf{I}_{gt}).
\end{equation}

The forward diffusion process adds noise to the ground truth latent representation $\mathbf{Z}^{gt}$ according to the DDIM schedule:

\begin{equation}
\mathbf{Z}_t^{enh} = \sqrt{\bar{\alpha}_t}\mathbf{Z}^{gt,enh} + \sqrt{1-\bar{\alpha}_t}\epsilon, \quad
\epsilon \sim \mathcal{N}(\mathbf{0}, \mathbf{I}).
\end{equation}

The reverse process employs a conditional UNet network $\epsilon_\theta^{enh}$ to predict the noise, conditioned on the LQ latent $\mathbf{Z}_{lq}$ and SII guidance:

\begin{equation}
\epsilon_\theta^{enh} = \text{UNet}^{enh}(\mathbf{Z}_t^{enh}, t, \mathbf{Z}_{lq}, \mathbf{F}_{oi}^{(t)}).
\end{equation}
where $t$ is the timestep, and $\mathbf{F}_{oi}^{(t)}$ is the guided feature from the SII module at step $t$.

During inference, given only an LQ image $\mathbf{X}^{lq}$, we encode it to obtain $\mathbf{Z}_{lq}$. We then initialize with random noise $\mathbf{Z}_T^{enh} \sim \mathcal{N}(0, \mathbf{I})$ and perform DDIM sampling:

\begin{equation}
\begin{split}
\mathbf{Z}_{t-1}^{\text{enh}} 
= \sqrt{\bar{\alpha}_{t-1}}\left(
\frac{\mathbf{Z}_t^{\text{enh}} - \sqrt{1-\bar{\alpha}_t}\epsilon_\theta^{\text{enh}}}
{\sqrt{\bar{\alpha}_t}}
\right) \\
+ \sqrt{1-\bar{\alpha}_{t-1} - \sigma_t^2}\epsilon_\theta^{\text{enh}} 
+ \sigma_t\epsilon_t.
\end{split}
\end{equation}

where $\sigma_t = \eta\sqrt{(1-\bar{\alpha}_{t-1})/(1-\bar{\alpha}_t)}\sqrt{1-\bar{\alpha}_t/\bar{\alpha}_{t-1}}$ and $\eta\in [0,1]$ controls stochasticity.

The final enhanced image is obtained by decoding the denoised latent:
\begin{equation}
\mathbf{I}_p = \text{Decoder}(\mathbf{Z}_0^{enh}).
\end{equation}

In the DiSeg branch, we employ DDIM for efficient segmentation mask generation.

During training, given an LQ image $\mathbf{X}^{lq}$ and its ground truth segmentation mask $\mathbf{S}_{gt}$, we encode the segmentation mask:

\begin{equation}
\mathbf{Z}^{gt,seg} = \text{Encoder}^{seg}(\mathbf{S}_{gt}).
\end{equation}

The forward diffusion process adds noise to the ground truth segmentation latent:

\begin{equation}
\mathbf{Z}_t^{seg} = \sqrt{\bar{\alpha}_t}\mathbf{Z}^{gt,seg} + \sqrt{1-\bar{\alpha}_t}\epsilon, \quad \epsilon \sim \mathcal{N}(\mathbf{0}, \mathbf{I}).
\end{equation}

The segmentation denoiser $\epsilon_\theta^{seg}$ predicts noise conditioned on LQ image features and SII guidance:

\begin{equation}
\mathbf{F}_{lq} = \text{Seg Encoder}(\mathbf{X}^{lq}).
\end{equation}
\begin{equation}
\epsilon_\theta^{seg} = \text{UNet}^{seg}(\mathbf{Z}_t^{seg}, t, \mathbf{F}_{lq}, \mathbf{F}_{os}^{(t)}).
\end{equation}
where $\mathbf{F}_{lq}$ are LQ image features and $\mathbf{F}_{os}^{(t)}$ is the guided feature from SII.

During inference, we initialize with random noise $\mathbf{Z}_T^{seg} \sim \mathcal{N}(0, \mathbf{I})$ and perform DDIM sampling:

\begin{equation}
\begin{split}
\mathbf{Z}_{t-1}^{\text{seg}} 
&= \sqrt{\bar{\alpha}_{t-1}}
\biggl(
\frac{\mathbf{Z}_t^{\text{seg}} - \sqrt{1-\bar{\alpha}_t}\,\epsilon_\theta^{\text{seg}}}
{\sqrt{\bar{\alpha}_t}}
\biggr) \\
&\quad + \sqrt{1-\bar{\alpha}_{t-1} - \sigma_t^2}\,\epsilon_\theta^{\text{seg}} 
+ \sigma_t\epsilon_t.
\end{split}
\end{equation}

The final segmentation mask is obtained by decoding:
\begin{equation}
\mathbf{S}_p = \text{Sigmoid}(\text{Decoder}^{seg}(\mathbf{Z}_0^{seg})).
\end{equation}

At each reverse step $t$ during the DDIM sampling process, $\mathbf{Z}_t^{enh}$ and $\mathbf{Z}_t^{seg}$ are processed through the SII module. The SII generates guided features $\mathbf{F}_{oi}^{(t)}$ for the DiEnh branch and $\mathbf{F}_{os}^{(t)}$ for the DiSeg branch, enabling mutual information exchange. The DDIM framework allows for flexible scheduling of the SII interactions. We employ a strategy where SII is applied at key steps $\tau \subset \{1, 2, ..., T\}$, with denser interactions at critical denoising stages.

\begin{figure}[t]
\centering
\includegraphics[scale=0.28]{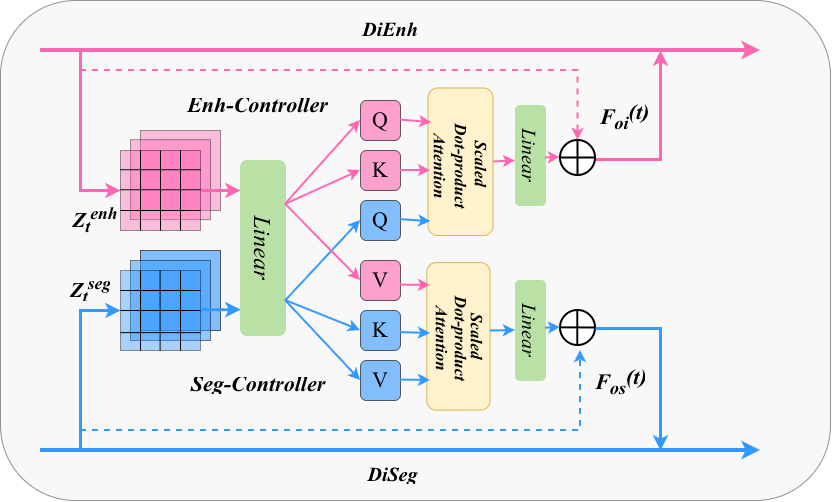}
\caption{Diagram illustrating the Symbiotic Information interaction module within the DDIM framework. Enh-Controller treats $\mathbf{Z}_t^{seg}$ as the query matrix and $\mathbf{Z}_t^{enh}$ as the key and value matrices to compute cross-attention, generating refined image feature $\mathbf{F}_{oi}^{(t)}$ for DiEnh branch. The Seg-Controller treats $\mathbf{Z}_t^{enh}$ as the query matrix and $\mathbf{Z}_t^{seg}$ as the key and value matrices to compute cross-attention, generating guided feature $\mathbf{F}_{os}^{(t)}$ for DiSeg branch.} \label{fig2}
\end{figure}

\subsection{Symbiotic Information Interaction Module}

To achieve information interaction between the DiEnh and DiSeg branches within the DDIM framework, we propose a Symbiotic Information Interaction (SII) module for joint optimization. SII consists of Enh-Controller and Seg-Controller, as shown in Fig.\ref{fig2}.

At time step $t$ during the DDIM reverse process, the Enh-Controller extracts updated image information from $\mathbf{Z}_t^{enh}$ in DiEnh and segmentation information from $\mathbf{Z}_t^{seg}$, generating a guided feature $\mathbf{F}_{oi}^{(t)}$ for DiEnh. Specifically, the Enh-Controller adopts the Multi-Head Attention mechanism to compute refined image features:

\begin{equation}
\mathbf{F}_{oic}^{(t)} = \text{MultiHeadAttention}\left(\mathbf{Z}_t^{seg}, \mathbf{Z}_t^{enh}, \mathbf{Z}_t^{enh}\right),
\end{equation}
where $\mathbf{Z}_t^{seg}$ serves as the query, and $\mathbf{Z}_t^{enh}$ serves as both key and value matrices.

We then apply $\operatorname{ReLU}(\cdot)$ to the enhanced image feature and combine it with the input image feature:

\begin{equation}
\mathbf{F}_{oi}^{(t)} = \operatorname{ReLU}\left(\mathbf{F}_{oic}^{(t)}\right) + \mathbf{Z}_t^{enh}.
\end{equation}

The Seg-Controller extracts updated segmentation information from $\mathbf{Z}_t^{seg}$ in DiSeg and image information from $\mathbf{Z}_t^{enh}$, generating a guided feature $\mathbf{F}_{os}^{(t)}$ for DiSeg:

\begin{equation}
\mathbf{F}_{osc}^{(t)} = \text{MultiHeadAttention}\left(\mathbf{Z}_t^{enh}, \mathbf{Z}_t^{seg}, \mathbf{Z}_t^{seg}\right),
\end{equation}
where $\mathbf{Z}_t^{enh}$ serves as the query, and $\mathbf{Z}_t^{seg}$ serves as key and value matrices.

Then, we apply $\operatorname{ReLU}(\cdot)$ to the segmentation feature and add it to the input segmentation feature:

\begin{equation}
\mathbf{F}_{os}^{(t)} = \operatorname{ReLU}\left(\mathbf{F}_{osc}^{(t)}\right) + \mathbf{Z}_t^{seg}.
\end{equation}

The final image feature $\mathbf{F}_{oi}^{(t)}$ and segmentation feature $\mathbf{F}_{os}^{(t)}$ are then used as conditioning inputs for the next DDIM sampling step.

\subsection{Joint Training Mechanism}

For DiSIINet, we extend the DDIM training objective to incorporate symbiotic interactions:

\begin{equation}
\mathcal{L}_{\text{DiEnh}}^{\text{DDIM}} = \mathbb{E}_{t,\mathbf{z}_0^{enh},\epsilon}\left[\|\epsilon - \epsilon_\theta^{enh}(\mathbf{Z}_t^{enh}, t, \mathbf{Z}_{lq}, \mathbf{F}_{oi}^{(t)})\|^2\right].
\end{equation}

\begin{equation}
\mathcal{L}_{\text{DiSeg}}^{\text{DDIM}} = \mathbb{E}_{t,\mathbf{z}_0^{seg},\epsilon}\left[\|\epsilon - \epsilon_\theta^{seg}(\mathbf{Z}_t^{seg}, t, \mathbf{F}_{lq}, \mathbf{F}_{os}^{(t)})\|^2\right].
\end{equation}

The final enhanced image $\mathbf{I}_p$ from DiEnh is supervised by a restoration loss:

\begin{equation}
\mathcal{L}_{\text{enh}} = \|\mathbf{I}_p - \mathbf{I}_{gt}\|^2.
\end{equation}
where $\mathbf{I}_{gt}$ is the ground truth high-quality image.

The final segmentation mask $\mathbf{S}_p$ from DiSeg is optimized using a weighted binary cross-entropy loss:

\begin{equation}
\mathcal{L}_{\text{seg}} = -\frac{1}{N} \sum w_i\left[\mathbf{S}_{gt} \log (\mathbf{S}_p) + (1-\mathbf{S}_{gt}) \log (1-\mathbf{S}_p)\right].
\end{equation}
where $\mathbf{S}_{gt}$ is the ground truth mask.

The overall loss integrates both DDIM training losses and final output supervision:

\begin{equation}
\mathcal{L}_{\text{overall}} = \mathcal{L}_{\text{DiEnh}}^{\text{DDIM}} + \beta\mathcal{L}_{\text{DiSeg}}^{\text{DDIM}} + \mathcal{L}_{\text{enh}} + \lambda\mathcal{L}_{\text{seg}}.
\label{equ8}
\end{equation}
where $\beta$ balances the DDIM training losses between branches, and $\lambda$ weights the segmentation loss relative to enhancement loss.

\section{Experiments and Results}

\subsection{Datasets}

To evaluate our DiSIINet model's performance, we conduct experiments on several medical image segmentation datasets including ACDC ~\cite{8360453}, KiTS19 ~\cite{heller2021state}, and TN3K ~\cite{gong2021multi}. ACDC dataset is a heart dataset featuring segmentation annotations for the the LV and RV volumes and the MYO. ACDC contains 150 patients, 100 training patients, and 50 test patients. The publicly accessible KiTS19 dataset comprises CT data from 300 patients. Specifically, cases indexed from 0 to 209 constitute the training set, while cases from 210 to 299 are designated as the test set. The TN3K (Thyroid Nodule Region Segmentation Dataset) consists of 3493 ultrasound images from 2421 patients. TN3K dataset contains 2879 images for training and 614 images for testing. For low-quality and high-quality image pairs, we use the original image as high-quality images, and we degraded the image as STS-SR ~\cite{11010915} method through down sampling to obtain low-quality images.

\subsection{Experimental Setup}

All experiments were performed using the PyTorch framework and executed on 4 NVIDIA GeForce RTX 3090 GPUs. For 3D datasets like MRI ACDC or CT KiTS19, we split them into 2D with a slice depth of one and corresponding labels were resized to dimensions of 320$\times$320 pixels. In contrast, images from the TN3K datasets were uniformly resized to a resolution of 256$\times$256 pixels. The networks were trained in an end-to-end manner utilizing the AdamW optimizer with a batch size of 32. The starting learning rate was established at 1$\times$$10^{-4}$. We conducted a joint loss function was employed to optimize the network.

\subsection{Implementation Details}

Our DiSIINet follows the standard DDIM framework: the model is trained with the full $T=1000$ diffusion steps (as in DDPM) to learn the complete denoising process, while during inference we adopt DDIM's accelerated sampling with only $S=50$ steps. This is achieved by selecting a subsequence of timesteps $\{\tau_1, \tau_2, ..., \tau_S\}$ where $\tau_S = T$, allowing efficient generation while preserving distribution fidelity. We employ a cosine noise schedule \cite{nichol2021improved} for $\bar{\alpha}_t$:

\begin{equation}
\bar{\alpha}_t = \frac{\cos(\pi t/2T + s)}{1 + s}, \quad s = 0.008
\end{equation}

The SII module is applied at all sampling steps to ensure continuous information exchange. For training, we use $\eta=1$ (fully stochastic) to improve robustness, while for inference we use $\eta=0$ (deterministic) for reproducibility. The weighting factors are set through ablation studies: $\beta=1.0$ to equally weight both DDIM training losses.

\begin{table*}[!h]
\centering
\scalebox{1.0}{
\begin{tabular}{c|c c|c c|c c}
\hline 
\multirow{2}{*}{Method} & \multicolumn{2}{|c|}{ACDC} & \multicolumn{2}{|c|}{KiTS19} & \multicolumn{2}{|c}{TN3K} \\ 
\cline{2-7} 
 & PSNR & SSIM & PSNR & SSIM & PSNR & SSIM \\ 
\hline 
PCENet ~\cite{liu2022degradation} & $19.86_{\textcolor{red}{\downarrow 12.29}}$& $0.832_{\textcolor{red}{\downarrow 0.100}}$ & $23.63_{\textcolor{red}{\downarrow 11.05}}$ & $0.703_{\textcolor{red}{\downarrow 0.258}}$ & $24.78_{\textcolor{red}{\downarrow 9.48}}$ & $0.894_{\textcolor{red}{\downarrow 0.048}}$\\ 

ArcNet ~\cite{li2022annotation} & $22.85_{\textcolor{red}{\downarrow 9.30}}$ & $0.824_{\textcolor{red}{\downarrow 0.108}}$ & $22.05_{\textcolor{red}{\downarrow 12.63}}$ & $0.802_{\textcolor{red}{\downarrow 0.159}}$ & $19.88_{\textcolor{red}{\downarrow 14.38}}$ & $0.851_{\textcolor{red}{\downarrow 0.091}}$  \\ 

SCRNet ~\cite{li2022structure} & $24.05_{\textcolor{red}{\downarrow 8.10}}$ & $0.886_{\textcolor{red}{\downarrow 0.046}}$ & $23.46_{\textcolor{red}{\downarrow 11.22}}$ & $0.854_{\textcolor{red}{\downarrow 0.107}}$ & $23.62_{\textcolor{red}{\downarrow 10.64}}$ & $0.895_{\textcolor{red}{\downarrow 0.047}}$  \\ 

DeepRFT ~\cite{mao2023intriguing} & $26.35_{\textcolor{red}{\downarrow 5.80}}$ & $0.902_{\textcolor{red}{\downarrow 0.030}}$ & $30.45_{\textcolor{red}{\downarrow 4.23}}$ & $0.947_{\textcolor{red}{\downarrow 0.014}}$ & $26.44_{\textcolor{red}{\downarrow 7.82}}$ & $0.924_{\textcolor{red}{\downarrow 0.018}}$  \\ 

I-SECRET ~\cite{cheng2021secret} & $27.18_{\textcolor{red}{\downarrow 4.97}}$ & $0.896_{\textcolor{red}{\downarrow 0.036}}$ & $32.60_{\textcolor{red}{\downarrow 2.08}}$ & $0.934_{\textcolor{red}{\downarrow 0.027}}$ & $29.04_{\textcolor{red}{\downarrow 5.22}}$ & $0.898_{\textcolor{red}{\downarrow 0.044}}$  \\ 

CHLNet ~\cite{wang2024clinical} & $30.41_{\textcolor{red}{\downarrow 1.74}}$ & $0.921_{\textcolor{red}{\downarrow 0.011}}$ & $\underline{33.65}_{\textcolor{red}{\downarrow 1.03}}$ & $\underline{0.953}_{\textcolor{red}{\downarrow 0.008}}$ & $34.11_{\textcolor{red}{\downarrow 0.15}}$ & $0.929_{\textcolor{red}{\downarrow 0.013}}$ \\ 

StillGAN ~\cite{ma2021structure} & $24.97_{\textcolor{red}{\downarrow 7.18}}$ & $0.845_{\textcolor{red}{\downarrow 0.087}}$ & $27.34_{\textcolor{red}{\downarrow 7.34}}$ & $0.912_{\textcolor{red}{\downarrow 0.049}}$ & $25.89_{\textcolor{red}{\downarrow 8.87}}$ & $0.885_{\textcolor{red}{\downarrow 0.057}}$  \\ 

LED ~\cite{cheng2023learning} & $25.88_{\textcolor{red}{\downarrow 6.27}}$ & $0.918_{\textcolor{red}{\downarrow 0.014}}$ & $33.26_{\textcolor{red}{\downarrow 1.42}}$ & $0.923_{\textcolor{red}{\downarrow 0.038}}$ & $27.58_{\textcolor{red}{\downarrow 6.68}}$ & $0.846_{\textcolor{red}{\downarrow 0.096}}$  \\ 

Fast-DDPM ~\cite{jiang2025fast} & $30.65_{\textcolor{red}{\downarrow 1.50}}$ & $\underline{0.922}_{\textcolor{red}{\downarrow 0.010}}$ & $31.25_{\textcolor{red}{\downarrow 3.43}}$ & $0.948_{\textcolor{red}{\downarrow 0.013}}$ & $\underline{34.12}_{\textcolor{red}{\downarrow 0.14}}$ & $\underline{0.935}_{\textcolor{red}{\downarrow 0.007}}$ \\

Wang ~\cite{wang2024joint} & $\underline{30.81}_{\textcolor{red}{\downarrow 1.34}}$ & $0.916_{\textcolor{red}{\downarrow 0.016}}$ & $32.87_{\textcolor{red}{\downarrow 1.81}}$ & $0.936_{\textcolor{red}{\downarrow 0.025}}$ & $33.88_{\textcolor{red}{\downarrow 0.38}}$ & $0.926_{\textcolor{red}{\downarrow 0.016}}$ \\ 
\hline 
DiSIINet & $\mathbf{32.15}$ & $\mathbf{0.932}$ & $\mathbf{34.68}$ & $\mathbf{0.961}$ & $\mathbf{34.26}$ & $\mathbf{0.942}$ \\ 
\hline
\end{tabular}}
\renewcommand{\tablename}{Table}
\caption{\normalfont\normalsize Joint multi-task learning improves image enhancement Model comparison performance.}
\label{tab:Com1}
\end{table*}

\begin{figure*}[!h]
\centering
\includegraphics[scale=0.2]{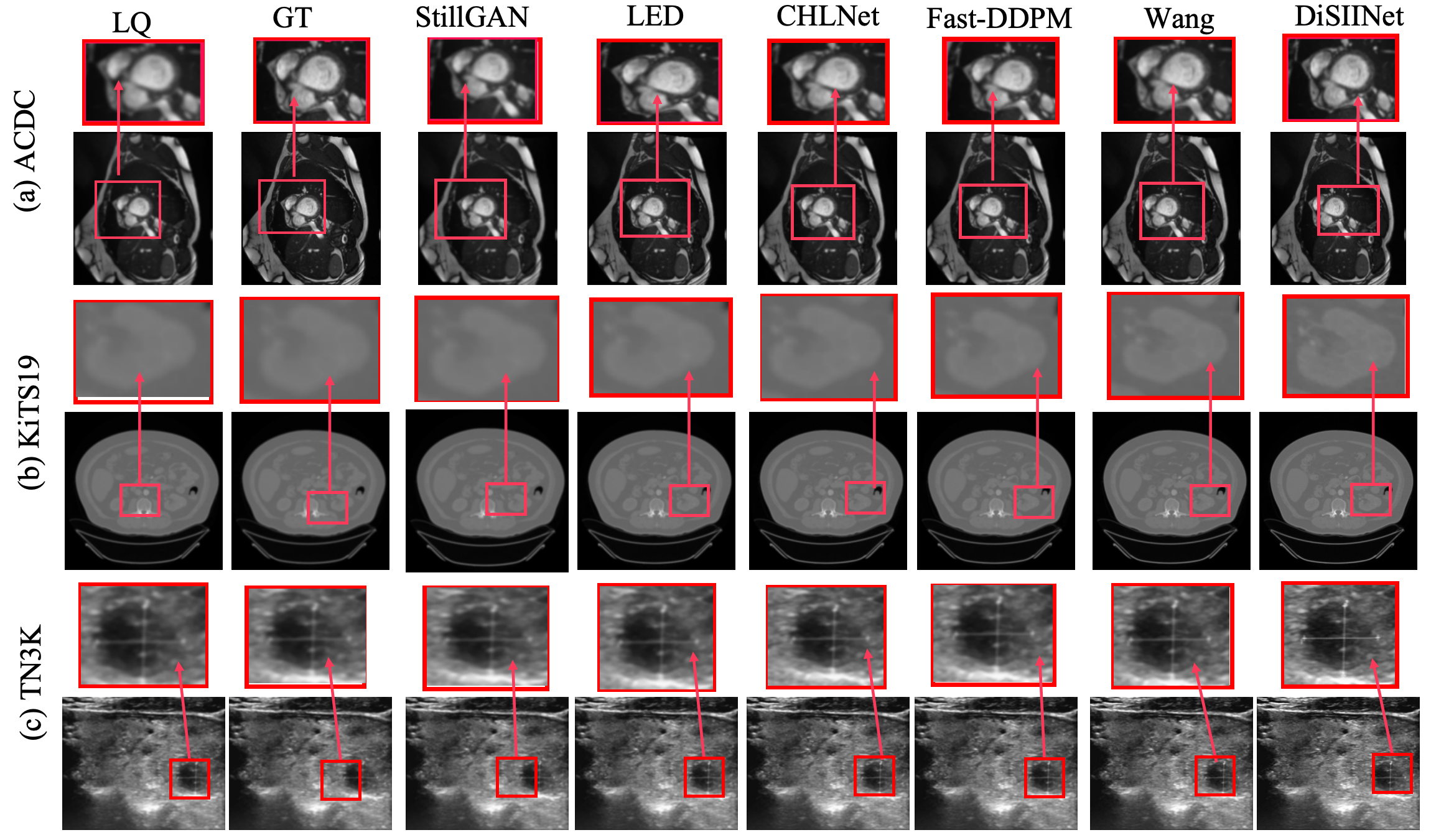}
\caption{Medical image enhancement task visual comparison results on three datasets. Example (a) row correspond to MRI ACDC images, example (b) row correspond to CT KiTS dataset and example (c) row correspond to ultrasound TN3K dataset.} \label{fig3}
\end{figure*}

\begin{table*}[!h]
\centering
\scalebox{1.0}{
\begin{tabular}{c|c c|c c|c c}
\hline 
\multirow{2}{*}{Method} & \multicolumn{2}{|c|}{ACDC} & \multicolumn{2}{|c|}{KiTS19} & \multicolumn{2}{|c}{TN3K} \\ 
\cline{2-7}
 & avg-mIoU & avg-Dice & avg-mIoU & avg-Dice & mIoU & Dice \\ 
\hline 
UNet~\cite{ronneberger2015u} & $70.07_{\textcolor{red}{\downarrow 15.05}}$ & $82.87_{\textcolor{red}{\downarrow 8.96}}$ & $59.60_{\textcolor{red}{\downarrow 17.65}}$ & $74.68_{\textcolor{red}{\downarrow 11.19}}$ & $64.20_{\textcolor{red}{\downarrow 10.98}}$ & $78.20_{\textcolor{red}{\downarrow 7.62}}$ \\ 

nnUNet~\cite{isensee2021nnu} & $82.26_{\textcolor{red}{\downarrow 2.55}}$ & $91.40_{\textcolor{red}{\downarrow 0.38}}$ & $68.40_{\textcolor{red}{\downarrow 6.85}}$ & $81.20_{\textcolor{red}{\downarrow 4.67}}$ & $66.10_{\textcolor{red}{\downarrow 9.08}}$ & $79.60_{\textcolor{red}{\downarrow 6.22}}$ \\ 

ResU-Net++~\cite{cao2022swin} & $75.00_{\textcolor{red}{\downarrow 9.81}}$ & $85.73_{\textcolor{red}{\downarrow 6.05}}$ & $64.20_{\textcolor{red}{\downarrow 11.05}}$ & $78.20_{\textcolor{red}{\downarrow 7.67}}$ & $67.80_{\textcolor{red}{\downarrow 7.38}}$ & $80.90_{\textcolor{red}{\downarrow 4.92}}$ \\ 

TransUNet~\cite{jha2021comprehensive} & $77.60_{\textcolor{red}{\downarrow 7.21}}$ & $87.46_{\textcolor{red}{\downarrow 4.32}}$ & $66.80_{\textcolor{red}{\downarrow 8.45}}$ & $80.16_{\textcolor{red}{\downarrow 5.71}}$ & $69.80_{\textcolor{red}{\downarrow 5.38}}$ & $82.16_{\textcolor{red}{\downarrow 3.66}}$ \\ 

SwinUNet~\cite{chen2021transunet} & $80.20_{\textcolor{red}{\downarrow 4.61}}$ & $89.15_{\textcolor{red}{\downarrow 2.63}}$ & $68.60_{\textcolor{red}{\downarrow 6.65}}$ & $81.38_{\textcolor{red}{\downarrow 4.49}}$ & $70.00_{\textcolor{red}{\downarrow 5.18}}$ & $82.40_{\textcolor{red}{\downarrow 3.42}}$ \\ 

DenoiSeg~\cite{buchholz2020denoiseg} & $83.27_{\textcolor{red}{\downarrow 1.54}}$ & $90.15_{\textcolor{red}{\downarrow 1.63}}$ & $67.80_{\textcolor{red}{\downarrow 7.45}}$ & $80.88_{\textcolor{red}{\downarrow 4.99}}$ & $70.40_{\textcolor{red}{\downarrow 4.78}}$ & $82.60_{\textcolor{red}{\downarrow 3.22}}$ \\ 

MedSegDiff-V2~\cite{wu2024medsegdiff} & $\underline{84.68}_{\textcolor{red}{\downarrow 0.13}}$ & $\underline{91.70}_{\textcolor{red}{\downarrow 0.08}}$ & $\underline{75.20}_{\textcolor{red}{\downarrow 0.05}}$ & $\underline{85.84}_{\textcolor{red}{\downarrow 0.03}}$ & $\underline{74.82}_{\textcolor{red}{\downarrow 0.36}}$ & $\underline{85.60}_{\textcolor{red}{\downarrow 0.22}}$ \\ 

Wang~\cite{wang2024joint} & $82.40_{\textcolor{red}{\downarrow 2.41}}$ & $90.33_{\textcolor{red}{\downarrow 1.45}}$ & $73.70_{\textcolor{red}{\downarrow 1.55}}$ & $84.78_{\textcolor{red}{\downarrow 1.09}}$ & $73.30_{\textcolor{red}{\downarrow 1.95}}$ & $84.60_{\textcolor{red}{\downarrow 1.22}}$ \\ 
\hline 
DiSIINet & $\mathbf{84.81}$ & $\mathbf{91.78}$ & $\mathbf{75.25}$ & $\mathbf{85.87}$ & $\mathbf{75.18}$ & $\mathbf{85.82}$ \\ 
\hline
\end{tabular}}
\renewcommand{\tablename}{Table}
\caption{\normalfont\normalsize Joint multi-task learning improves image segmentation model comparison performance.} 
\label{tab:Com2}
\end{table*}

\begin{figure*}[h]
\centering
\includegraphics[scale=0.26]{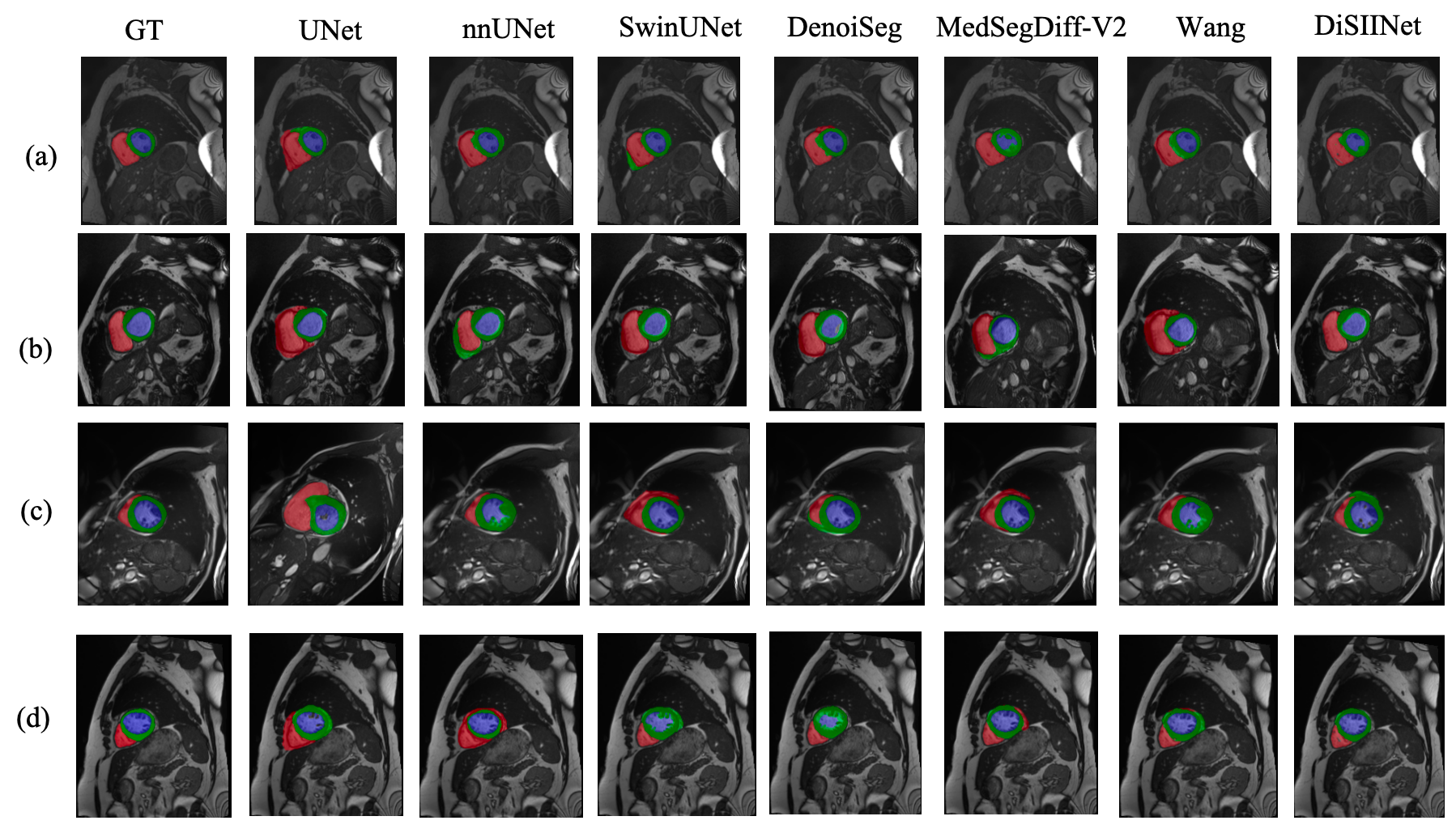}
\caption{Medical image segmentation task visual comparison results on MRI ACDC dataset.} \label{fig4}
\vspace{-1em}
\end{figure*}

\subsection{Joint multi-task learning improves medical image enhancement}

The quantitative results on medical image enhancement task are shown in Table \ref{tab:Com1}. Among the methods compared on ACDC datasets, DiSIINet demonstrates superior performance, achieving a PSNR of 32.15 and an SSIM of 0.932. This performance positions DiSIINet at the forefront of the methodologies evaluated in the ACDC dataset. Figure.~\ref{fig3} example (a) presents a qualitative comparison for medical image enhancement using the MRI ACDC dataset. The enlarged area, emphasized by red rectangular boxes, illustrates DiSIINet's effectiveness in maintaining fine structural details in the denoised images relative to other techniques. Remarkably, the ventricular region is depicted with the greatest clarity and sharpness in the DiSIINet output, while competing methods demonstrate significant blurring and loss of detail in this area. The pulmonary fissure serves as the boundary between the ventricular lobes. Within the methods evaluated on the KiTS19 datasets, DiSIINet achieves a PSNR of 34.68 and an SSIM of 0.961, marking it as the leading approach for this dataset. These results not only surpass the performance of other methods but also indicate a superior ability to maintain both pixel accuracy and perceptual quality in the enhanced images. Figure.~\ref{fig3} example (b) presents a qualitative comparison for medical image enhancement using the CT KiTS19 dataset. The enlarged red area contains the kidney and kidney tumor. The kidney organ and tumor are distinctly shown with contour boundaries in the DiSIINet enhanced images, whereas they cannot be differentiated by other methods. In the comparison of methods using the TN3K datasets, DiSIINet stands out as the top performer on the TN3K dataset, achieving a PSNR of 34.26 and an SSIM of 0.942. This remarkable performance underscores the model's effectiveness in enhancing image quality while preserving structural integrity, making it a benchmark in the field. Figure.~\ref{fig3} example (c) presents a qualitative comparison for medical image enhancement using the ultrasound TN3K. The enlarged red area shows the thyroid nodule, which has a clear contour against the surrounding background in the DiSIINet enhanced images. In contrast, it appears more blurred using other methods.

\subsection{Joint multi-task learning improves medical image segmentation}

In analyzing the performance of various methods on the ACDC dataset as presented in Table ~\ref{tab:Com2}, DiSIINet emerges as the top-performing model, achieving an avg-mIoU of 84.81 and an avg-Dice of 91.78. The strong results indicate that the model excels in capturing the intricate features necessary for effective segmentation. We extensively evaluated the deep models on MRI ACDC datsets with visual results, as shown in Figure.~\ref{fig4}. It is visible that the DiSIINet can produce clean medical images without any visually disturbing artifacts for denoising on ACDC dataset. We can observe the rich hierarchical structure and distinct boundaries among the different heart organizations, containing the LV, RV, and MYO. The tiny but vital details of the heart become gradually obscured. This phenomenon indicates that the proposed DiSIINet can capture the tiny but vital structures hidden under the noise even if the noise level is high, thanks to the ability to abundantly collect information and preserve intricate structures through SII modules.

In evaluating the performance of various methods on the KiTS19 dataset, it is evident that joint multi-task learning significantly enhances segmentation capabilities. DiSIINet emerges as the leading model, achieving an mIoU of 75.25 and a Dice coefficient of 85.87. These results not only position DiSIINet at the forefront of the evaluated methods but also demonstrate its effectiveness in accurately segmenting images within the KiTS19 dataset. The high performance indicates that DiSIINet is adept at addressing the intricate challenges associated with this type of imagery. 

In assessing the performance of different methods on the TN3K dataset, it is clear that joint multi-task learning significantly enhances segmentation outcomes, as evidenced by the metrics of mIoU and Dice coefficient. DiSIINet stands out as the top-performing model, achieving an mIoU of 75.18 and a Dice coefficient of 85.82. These exceptional results highlight DiSIINet's ability to effectively segment images within the TN3K dataset, demonstrating a robust performance that surpasses that of competing methodologies. The high Dice coefficient, in particular, indicates a strong alignment between the predicted and ground truth segmentation, reflecting the model's superior capability in capturing intricate details. 

\subsection{Ablation Study}

\subsubsection{Effectiveness of Key Component}

\begin{table}[h]
\centering
\scalebox{0.62}{
\begin{tabular}{c|c|c|c|c|c|c}
\hline 
\multirow{2}{*}{Method} & \multicolumn{2}{|c|}{ACDC} & \multicolumn{2}{|c|}{KiTS19} & \multicolumn{2}{|c}{TN3K} \\ 
\cline{2-7}
& PSNR & avg-Dice & PSNR & avg-Dice & PSNR & Dice \\ 
\hline 
\makecell{Cascaded} & 23.08 & 84.51 & 24.46 & 81.03 & 23.73 & 82.62 \\
\hline 
w/o Enh-Controller & 26.28 & 88.34 & 26.38 & 84.87 & 25.13 & 85.48 \\
\hline 
w/o Seg-Controller & 27.75 & 86.12 & 28.34 & 82.61 & 27.33 & 81.06 \\ 
\hline 
\makecell{Joint Enh-Controller \& \\ Seg-Controller (Ours DiSIINet)} & \textbf{32.15} & \textbf{91.83} & \textbf{34.68} & \textbf{86.89} & \textbf{34.26} & \textbf{86.62} \\ 
\hline 
\end{tabular}}
\renewcommand{\tablename}{Table}
\caption{\normalfont\normalsize Ablation study of each component of the SII module.}
\label{tab:Abl0}
\end{table}

Table~\ref{tab:Abl0} highlights the contributions of different components in DiSIINet. Our proposed DISIINet, which integrates both the Enh-Controller and Seg-Controller, achieves the highest performance across all datasets. For comparison, a baseline cascaded method where low-quality images are first enhanced with Fast-DDPM and then segmented using MedSegDiff-V2 exhibits a significant performance decline. This result underscores the critical contribution of each controller to optimal outcomes. Specifically, removing the Enh-Controller from DISIINet leads to decreased metrics. Conversely, removing the Seg-Controller results in even lower PSNR values and a more pronounced decline in Dice coefficients. These ablation studies confirm that the synergistic integration of both controllers is essential for simultaneously maximizing image quality and segmentation accuracy, making DISIINet model the most effective configuration.

\subsubsection{Sensitivity analysis of loss weights}

\begin{table}[h]
\centering
\scalebox{0.75}{
\begin{tabular}{c|c|c|c|c|c|c}
\hline 
\multirow{2}{*}{$\lambda$} & \multicolumn{2}{|c|}{ACDC} & \multicolumn{2}{|c|}{KiTS19} & \multicolumn{2}{|c}{TN3K} \\ 
\cline{2-7}
 & PSNR & avg-Dice & PSNR & avg-Dice & PSNR & Dice \\ 
\hline 
$\lambda=3 \times 10^{-1}$ & 31.55 & 90.74 & 34.53 & 85.32 & 34.23 & 85.12 \\
\hline 
$\lambda=4 \times 10^{-1}$ & 31.26 & 91.39 & 34.31 & 85.68 & 34.18 & 85.36 \\
\hline 
$\lambda=5 \times 10^{-1}$ & \textbf{32.15} & \textbf{91.83} & \textbf{34.68} & \textbf{86.89} & \textbf{34.26} & \textbf{86.62} \\
\hline 
$\lambda=6 \times 10^{-1}$ & 30.62 & 91.58 & 33.26 & 86.18 & 33.89 & 86.33 \\
\hline
\end{tabular}}
\renewcommand{\tablename}{Table}
\caption{\normalfont\normalsize Ablation study of weighting parameter $\lambda$.}
\label{tab:Abl1}
\vspace{-1em}
\end{table}

Finally, we evaluate the impact of different weighting parameter $\lambda$ between the DiEnh and DiSeg branches during training on performance using three dataset. The choice of loss weighting parameter $\lambda$ is entirely result-driven. Table ~\ref{tab:Abl1} shows that our DiSIINet achieves an optimal result when $\lambda$=5$\times$ $10^{-1}$. Therefore, we use this setting in all experiments.

\section{Conclusion}
This paper presents DiSIINet, a dual-DDIM diffusion symbiotic network that joint medical image enhancement and segmentation. By coupling two diffusion processes via the Symbiotic Information Interaction (SII) module, DiSIINet enables continuous, feature-level information exchange that allows enhancement and segmentation tasks to mutually reinforce each other. Extensive experiments on multi-modal medical datasets demonstrate that DiSIINet yields superior performance compared to sequential or independent approaches. However, the performance gains come at the cost of increased computational complexity. Future work will focus on exploring distillation techniques to smaller models, and investigating adaptive mechanisms to reduce the required denoising steps without compromising the symbiotic benefits.

\vspace{-0.5em}
\section*{Acknowledgements}
This research is part of the IN-CYPHER programme and is supported by the National Research Foundation, Prime Minister's Office, Singapore under its Campus for Research Excellence and Technological Enterprise (CREATE) programme.

\bibliographystyle{named}
\bibliography{ijcai26}

@inproceedings{wang2024clinical,
  title={A Clinical-Oriented Lightweight Network for High-Resolution Medical Image Enhancement},
  author={Wang, Yaqi and Chen, Leqi and Hou, Qingshan and Cao, Peng and Yang, Jinzhu and Liu, Xiaoli and Zaiane, Osmar R},
  booktitle={International Conference on Medical Image Computing and Computer-Assisted Intervention},
  pages={3--12},
  year={2024},
  organization={Springer}
}

@inproceedings{liu2022degradation,
  title={Degradation-invariant enhancement of fundus images via pyramid constraint network},
  author={Liu, Haofeng and Li, Heng and Fu, Huazhu and Xiao, Ruoxiu and Gao, Yunshu and Hu, Yan and Liu, Jiang},
  booktitle={International Conference on Medical Image Computing and Computer-Assisted Intervention},
  pages={507--516},
  year={2022},
  organization={Springer}
}

@article{li2022annotation,
  title={An annotation-free restoration network for cataractous fundus images},
  author={Li, Heng and Liu, Haofeng and Hu, Yan and Fu, Huazhu and Zhao, Yitian and Miao, Hanpei and Liu, Jiang},
  journal={IEEE Transactions on Medical Imaging},
  volume={41},
  number={7},
  pages={1699--1710},
  year={2022},
  publisher={IEEE}
}

@inproceedings{li2022structure,
  title={Structure-consistent restoration network for cataract fundus image enhancement},
  author={Li, Heng and Liu, Haofeng and Fu, Huazhu and Shu, Hai and Zhao, Yitian and Luo, Xiaoling and Hu, Yan and Liu, Jiang},
  booktitle={International Conference on Medical Image Computing and Computer-Assisted Intervention},
  pages={487--496},
  year={2022},
  organization={Springer}
}

@article{ma2021structure,
  title={Structure and illumination constrained GAN for medical image enhancement},
  author={Ma, Yuhui and Liu, Jiang and Liu, Yonghuai and Fu, Huazhu and Hu, Yan and Cheng, Jun and Qi, Hong and Wu, Yufei and Zhang, Jiong and Zhao, Yitian},
  journal={IEEE Transactions on Medical Imaging},
  volume={40},
  number={12},
  pages={3955--3967},
  year={2021},
  publisher={IEEE}
}

@inproceedings{mao2023intriguing,
  title={Intriguing findings of frequency selection for image deblurring},
  author={Mao, Xintian and Liu, Yiming and Liu, Fengze and Li, Qingli and Shen, Wei and Wang, Yan},
  booktitle={Proceedings of the AAAI Conference on Artificial Intelligence},
  volume={37},
  number={2},
  pages={1905--1913},
  year={2023}
}

@inproceedings{cheng2021secret,
  title={I-secret: Importance-guided fundus image enhancement via semi-supervised contrastive constraining},
  author={Cheng, Pujin and Lin, Li and Huang, Yijin and Lyu, Junyan and Tang, Xiaoying},
  booktitle={Medical Image Computing and Computer Assisted Intervention--MICCAI 2021: 24th International Conference, Strasbourg, France, September 27--October 1, 2021, Proceedings, Part VIII 24},
  pages={87--96},
  year={2021},
  organization={Springer}
}

@article{cheng2023learning,
  title={Learning enhancement from degradation: A diffusion model for fundus image enhancement},
  author={Cheng, Puijin and Lin, Li and Huang, Yijin and He, Huaqing and Luo, Wenhan and Tang, Xiaoying},
  journal={arXiv preprint arXiv:2303.04603},
  year={2023}
}

@inproceedings{ronneberger2015u,
  title={U-net: Convolutional networks for biomedical image segmentation},
  author={Ronneberger, Olaf and Fischer, Philipp and Brox, Thomas},
  booktitle={MICCAI},
  year={2015},
  organization={Springer}
}

@article{jha2021comprehensive,
  title={A comprehensive study on colorectal polyp segmentation with ResUNet++, conditional random field and test-time augmentation},
  author={Jha, Debesh and Smedsrud, Pia H and Johansen, Dag and De Lange, Thomas and Johansen, H{\aa}vard D and Halvorsen, P{\aa}l and Riegler, Michael A},
  journal={IEEE Journal of Biomedical and Health Informatics},
  year={2021},
  publisher={IEEE}
}

@inproceedings{cao2022swin,
  title={Swin-unet: Unet-like pure transformer for medical image segmentation},
  author={Cao, Hu and Wang, Yueyue and Chen, Joy and Jiang, Dongsheng and Zhang, Xiaopeng and Tian, Qi and Wang, Manning},
  booktitle={ECCVW},
  year={2022},
  organization={Springer}
}

@article{chen2021transunet,
  title={Transunet: Transformers make strong encoders for medical image segmentation},
  author={Chen, Jieneng and Lu, Yongyi and Yu, Qihang and Luo, Xiangde and Adeli, Ehsan and Wang, Yan and Lu, Le and Yuille, Alan L and Zhou, Yuyin},
  journal={arXiv preprint arXiv:2102.04306},
  year={2021}
}

@inproceedings{wang2024joint,
  title={Joint EM Image Denoising and Segmentation with Instance-Aware Interaction},
  author={Wang, Zhicheng and Li, Jiacheng and Chen, Yinda and Shou, Jiateng and Deng, Shiyu and Huang, Wei and Xiong, Zhiwei},
  booktitle={International Conference on Medical Image Computing and Computer-Assisted Intervention},
  pages={403--413},
  year={2024},
  organization={Springer}
}

@inproceedings{rombach2022high,
  title={High-resolution image synthesis with latent diffusion models},
  author={Rombach, Robin and Blattmann, Andreas and Lorenz, Dominik and Esser, Patrick and Ommer, Bj{\"o}rn},
  booktitle={Proceedings of the IEEE/CVF conference on computer vision and pattern recognition},
  pages={10684--10695},
  year={2022}
}

@ARTICLE{8360453,
  author={Bernard, Olivier and Lalande, Alain and Zotti. et al.},
  journal={IEEE Transactions on Medical Imaging}, 
  title={Deep Learning Techniques for Automatic MRI Cardiac Multi-Structures Segmentation and Diagnosis: Is the Problem Solved?}, 
  year={2018},
  volume={37},
  number={11},
  pages={2514-2525},
  doi={10.1109/TMI.2018.2837502}}

@article{ho2020denoising,
  title={Denoising diffusion probabilistic models},
  author={Ho, Jonathan and Jain, Ajay and Abbeel, Pieter},
  journal={Advances in neural information processing systems},
  volume={33},
  pages={6840--6851},
  year={2020}
}

@inproceedings{liang2022efficient,
  title={Efficient and degradation-adaptive network for real-world image super-resolution},
  author={Liang, Jie and Zeng, Hui and Zhang, Lei},
  booktitle={European Conference on Computer Vision},
  pages={574--591},
  year={2022},
  organization={Springer}
}

@inproceedings{park2023content,
  title={Content-aware local gan for photo-realistic super-resolution},
  author={Park, JoonKyu and Son, Sanghyun and Lee, Kyoung Mu},
  booktitle={Proceedings of the IEEE/CVF International Conference on Computer Vision},
  pages={10585--10594},
  year={2023}
}

@inproceedings{wang2023exploring,
  title={Exploring clip for assessing the look and feel of images},
  author={Wang, Jianyi and Chan, Kelvin CK and Loy, Chen Change},
  booktitle={Proceedings of the AAAI conference on artificial intelligence},
  volume={37},
  number={2},
  pages={2555--2563},
  year={2023}
}

@inproceedings{lin2024diffbir,
  title={Diffbir: Toward blind image restoration with generative diffusion prior},
  author={Lin, Xinqi and He, Jingwen and Chen, Ziyan and Lyu, Zhaoyang and Dai, Bo and Yu, Fanghua and Qiao, Yu and Ouyang, Wanli and Dong, Chao},
  booktitle={European conference on computer vision},
  pages={430--448},
  year={2024},
  organization={Springer}
}

@inproceedings{yang2024pixel,
  title={Pixel-aware stable diffusion for realistic image super-resolution and personalized stylization},
  author={Yang, Tao and Wu, Rongyuan and Ren, Peiran and Xie, Xuansong and Zhang, Lei},
  booktitle={European conference on computer vision},
  pages={74--91},
  year={2024},
  organization={Springer}
}

@inproceedings{wu2024seesr,
  title={Seesr: Towards semantics-aware real-world image super-resolution},
  author={Wu, Rongyuan and Yang, Tao and Sun, Lingchen and Zhang, Zhengqiang and Li, Shuai and Zhang, Lei},
  booktitle={Proceedings of the IEEE/CVF conference on computer vision and pattern recognition},
  pages={25456--25467},
  year={2024}
}

@inproceedings{buchholz2020denoiseg,
  title={DenoiSeg: joint denoising and segmentation},
  author={Buchholz, Tim-Oliver and Prakash, Mangal and Schmidt, Deborah and Krull, Alexander and Jug, Florian},
  booktitle={European Conference on Computer Vision},
  pages={324--337},
  year={2020},
  organization={Springer}
}

@inproceedings{liu2017image,
  title={When image denoising meets high-level vision tasks: A deep learning approach},
  author={Liu, Ding and Wen, Bihan and Liu, Xianming and Wang, Zhangyang and Huang, Thomas S},
  booktitle={Proceedings of the International Joint Conference on Artificial Intelligence},
pages = {842–848},
numpages = {7},
  year={2018}
}

@article{xu2023synergy,
  title={Synergy between semantic segmentation and image denoising via alternate boosting},
  author={Xu, Shunxin and Sun, Ke and Liu, Dong and Xiong, Zhiwei and Zha, Zheng-Jun},
  journal={ACM Transactions on Multimedia Computing, Communications and Applications},
  volume={19},
  number={2},
  pages={1--23},
  year={2023},
  publisher={ACM New York, NY}
}

@article{jiang2025fast,
  title={Fast-DDPM: Fast denoising diffusion probabilistic models for medical image-to-image generation},
  author={Jiang, Hongxu and Imran, Muhammad and Zhang, Teng and Zhou, Yuyin and Liang, Muxuan and Gong, Kuang and Shao, Wei},
  journal={IEEE Journal of Biomedical and Health Informatics},
  year={2025},
  publisher={IEEE}
}

@inproceedings{wu2024medsegdiff,
  title={Medsegdiff-v2: Diffusion-based medical image segmentation with transformer},
  author={Wu, Junde and Ji, Wei and Fu, Huazhu and Xu, Min and Jin, Yueming and Xu, Yanwu},
  booktitle={Proceedings of the AAAI conference on artificial intelligence},
  volume={38},
  number={6},
  pages={6030--6038},
  year={2024}
}

@inproceedings{zhai2022scaling,
  title={Scaling vision transformers},
  author={Zhai, Xiaohua and Kolesnikov, Alexander and Houlsby, Neil and Beyer, Lucas},
  booktitle={Proceedings of the IEEE/CVF conference on computer vision and pattern recognition},
  pages={12104--12113},
  year={2022}
}

@article{gao2023corediff,
  title={CoreDiff: Contextual error-modulated generalized diffusion model for low-dose CT denoising and generalization},
  author={Gao, Qi and Li, Zilong and Zhang, Junping and Zhang, Yi and Shan, Hongming},
  journal={IEEE Transactions on Medical Imaging},
  volume={43},
  number={2},
  pages={745--759},
  year={2023},
  publisher={IEEE}
}

@inproceedings{xiang2023ddm,
  title={DDM $^2$: Self-supervised diffusion MRI denoising with generative diffusion models},
  author={Xiang, Tiange and Yurt, Mahmut and Syed, Ali B and Setsompop, Kawin and Chaudhari, Akshay},
  booktitle={The Eleventh International Conference on Learnin Representations.},
  year={2023}
}

@article{isensee2021nnu,
  title={nnU-Net: a self-configuring method for deep learning-based biomedical image segmentation},
  author={Isensee, Fabian and Jaeger, Paul F and Kohl, Simon AA and Petersen, Jens and Maier-Hein, Klaus H},
  journal={Nature methods},
  volume={18},
  number={2},
  pages={203--211},
  year={2021},
  publisher={Nature Publishing Group}
}

@article{heller2021state,
  title={The state of the art in kidney and kidney tumor segmentation in contrast-enhanced CT imaging: Results of the KiTS19 challenge},
  author={Heller, Nicholas and Isensee, Fabian and Maier-Hein, Klaus H and Hou, Xiaoshuai and Xie, Chunmei and Li, Fengyi and Nan, Yang and Mu, Guangrui and Lin, Zhiyong and Han, Miofei and others},
  journal={Medical image analysis},
  volume={67},
  pages={101821},
  year={2021},
  publisher={Elsevier}
}

@inproceedings{gong2021multi,
  title={Multi-task learning for thyroid nodule segmentation with thyroid region prior},
  author={Gong, Haifan and Chen, Guanqi and Wang, Ranran and Xie, Xiang and Mao, Mingzhi and Yu, Yizhou and Chen, Fei and Li, Guanbin},
  booktitle={2021 IEEE 18th international symposium on biomedical imaging (ISBI)},
  pages={257--261},
  year={2021},
  organization={IEEE}
}

@ARTICLE{11010915,
  author={Li, Yinghua and Hao, Weiao and Zeng, Hao and Wang, Longguang and Xu, Jian and Routray, Sidheswar and Jhaveri, Rutvij H. and Gadekallu, Thippa Reddy},
  journal={IEEE Journal of Biomedical and Health Informatics}, 
  title={Cross-Scale Texture Supplementation for Reference-based Medical Image Super-Resolution}, 
  year={2025},
  volume={},
  number={},
  pages={1-15},
  doi={10.1109/JBHI.2025.3572502}}

@inproceedings{song2021denoising,
  title={Denoising Diffusion Implicit Models},
  author={Song, Jiaming and Meng, Chenlin and Ermon, Stefano},
  booktitle={International Conference on Learning Representations (ICLR)},
  year={2021}
}

@inproceedings{nichol2021improved,
  title={Improved Denoising Diffusion Probabilistic Models},
  author={Nichol, Alex and Dhariwal, Prafulla},
  booktitle={International Conference on Machine Learning (ICML)},
  pages={8162--8171},
  year={2021},
  organization={PMLR}
}

@article{xiao2024semantic,
  title={Semantic segmentation prior for diffusion-based real-world super-resolution},
  author={Xiao, Jiahua and Zhang, Jiawei and Zou, Dongqing and Zhang, Xiaodan and Ren, Jimmy and Wei, Xing},
  journal={arXiv preprint arXiv:2412.02960},
  year={2024}
}

\end{document}